\documentclass[sigplan, 10pt, screen]{acmart}

\usepackage{graphicx}
\usepackage{multirow}
\usepackage[linesnumbered,ruled,vlined]{algorithm2e}
\usepackage{tabularx}
\usepackage{nicematrix,enumitem,booktabs}
\usepackage{listings}
\usepackage{xcolor}
\usepackage{float}

\usepackage[frozencache=true,cachedir=minted-cache]{minted}
\usepackage{color, colortbl}
\definecolor{LightCyan}{rgb}{0.88,1,1}
\definecolor{my}{rgb}{0.61569,0.76471,0.90196}

\definecolor{Gray}{gray}{0.9}

\usepackage{etoolbox}
\makeatletter
\AtBeginEnvironment{minted}{\dontdofcolorbox}
\def\dontdofcolorbox{\renewcommand\fcolorbox[4][]{##4}}
\makeatother

\newcommand\myVSpace[1][5pt]{\rule[\normalbaselineskip]{0pt}{#1}}

\AtBeginDocument{%
  }

\begin{document}


%% TODO: Review which is better
% \title{Performant Triton Code Generation on GPU for Dense Operations}
\title{ML-Triton, A Multi-Level Compilation and Language Extension to Triton GPU Programming}
%\title{Efficient Triton GPU Code Generation for Dense Operations}


\author{Dewei Wang}
\affiliation{%
  \institution{\textit{Intel Corporation}}
  \city{Shanghai}
  \country{China}}
\email{dewei.wang@intel.com}

\author{Wei Zhu}
\affiliation{%
  \institution{\textit{Intel Corporation}}
  \city{Shanghai}
  \country{China}}
\email{wei2.zhu@intel.com}

\author{Liyang Ling}
\affiliation{%
  \institution{\textit{Intel Corporation}}
  \city{Shanghai}
  \country{China}}
\email{liyang.ling@intel.com}

\author{Ettore Tiotto}
\affiliation{%
  \institution{\textit{Intel Corporation}}
  \city{Toronto}
  \country{Canada}}
\email{ettore.tiotto@intel.com}

\author{Quintin Wang}
\affiliation{%
  \institution{\textit{Intel Corporation}}
  \city{Shanghai}
  \country{China}}
\email{quintin.wang@intel.com}

\author{Whitney Tsang}
\affiliation{%
  \institution{\textit{Intel Corporation}}
  \city{Toronto}
  \country{Canada}}
\email{whitney.tsang@intel.com}

\author{Julian Oppermann}
\affiliation{%
  \institution{\textit{Codeplay Software}}
  \city{London}
  \country{United Kingdom}}
\email{julian.oppermann@codeplay.com}

\author{Jacky Deng}
\affiliation{%
  \institution{\textit{Intel Corporation}}
  \city{Shanghai}
  \country{China}}
\email{jacky.deng@intel.com}


%\renewcommand{\shortauthors}{Dewei et al.}

\begin{abstract}
In the era of Large Language Models (LLMs), dense operations such as General Matrix Multiplication (GEMM) and Multi-Head Attention (MHA) are critical components. These operations are well-suited for parallel execution using a tile-based approach. While traditional GPU programming often relies on low level interfaces like CUDA or SYCL, Triton~\cite{triton} has emerged as a domain-specific language (DSL) that offers a more user-friendly and portable alternative by programming at a higher level.
%thread level?

The current Triton starts at the workgroup (aka threadblock) level, and directly lowers to per-thread level. And then attempt to coalesce and amend through a series of passes, promoting information from low-level representation. We believe this is pre-mature lowering based on the below observations.

\begin{enumerate}
    \item GPU has a hierarchical structure both physically and logically.  Modern GPUs often feature SIMD units capable of directly operating on tiles on a warp or warp-group basis, such as blocked load and blocked matrix multiply-accumulate (MMA).
    \item Multi-level gradual lowering can make compiler decoupled and clean by separating considerations inter and intra a logical layer. 
    %GPUs communicate data between warps via shared memory. intra-warp communication is done 
    %it is common that GPUs communicate data between warps via shared memory
    \item Kernel developers often need fine control to get good performance on the latest hardware. FlashAttention2~\cite{dao2023flashattention2fasterattentionbetter} advocates explicit data partition between warps to make a performance boost.
\end{enumerate}	

In this context, we propose ML-Triton which features multi-level compilation flow and programming interface. Our approach begins at the workgroup level and progressively lowers to the warp and intrinsic level, implementing a multi-level lowering align with the hierarchical nature of GPU. Additionally, we extend triton language to support user-set compiler hint and warp level programming, enabling researchers to get good out-of-the box performance without awaiting compiler updates. 

Experimental results demonstrate that our approach achieves performance above $95\%$ of expert-written kernels on Intel GPU, as measured by the geometric mean.
\end{abstract}

%    <concept>
%        <concept_id>10011007.10011006.10011041.10011047</concept_id>
%        <concept_desc>Software and its engineering~Source code generation</concept_desc>
%        <concept_significance>500</concept_significance>
%        </concept>
%  </ccs2012>

% \ccsdesc[500]{Software and its engineering~Source code generation}





\maketitle

\pagestyle{plain}

\section{Background}
\subsection{Intel GPU}
Released in 2022, the Intel Ponte Vecchio GPU (PVC)~\cite{pvc} is architected with a strong emphasis on AI inference and training workloads. This GPU features a modular design, consisting of two tiles, each containing 64 XeCores. Each XeCore houses 8 Execution Units (EUs), with each EU capable of supporting 8 hardware contexts, where each context operates using native SIMD16 instructions. This architecture positions PVC as a formidable General-Purpose GPU (GPGPU), delivering robust performance and efficiency across a broad range of applications.

PVC boasts an advanced memory hierarchy that significantly enhances efficiency and speed. The global memory offers extensive storage accessible by all XeCores within the GPU. The L2 cache acts as a high-speed intermediary, bridging global memory and core processing units to reduce latency and optimize overall performance. Additionally, the Shared Local Memory (SLM) and L1 cache, which share the same physical space, facilitate rapid data communication between warps, further optimizing performance. In terms of logical hierarchy, Intel employs the concepts of \emph{Workgroup}, \emph{Subgroup}, \emph{Workitem}, which are the counterpart of Nvidia \emph{CTA/ThreadBlock}, \emph{Warp}, \emph{Thread} respectively. The specification of PVC (max 1550) is summarized in Table~\ref{tab:pvc}.
%Since triton adopts Nvidia terminology, we repect this for ease of illustration.

Intel complements the hardware with a comprehensive instruction set that maximizes the GPU's potential. Recognizing the importance of memory bandwidth, PVC features 1D/2D blocked load/store/prefetch instructions, which enhances data transfer efficiency from HBM. These instructions support cache hint, address calculation offload, hardware data padding and hardware transpose. To accelerate compute-intensive tasks, PVC also has a DPAS (Dot Product Accumulation with Systolic Array) instruction which accelerates GEMM on blocked data.

PVC supports both SIMT and SIMD programming models, offering flexibility to developers. For SIMT, Intel promotes SYCL, a cross-platform C++ programming model with a thread-parallel execution approach closely aligned with CUDA. For SIMD, there is a SYCL extension called eSIMD~\cite{c-for-metal}, enables developers to write explicitly vectorized code.

\begin{table}
\caption{Hardware Specification for PVC max 1550}
\label{tab:pvc}
\small
\begin{tabularx}{\linewidth}{lllll}
\toprule
Hardware  & Logical     & Memory                            & \multirow{2}{*}{Capacity}         \\
Level     & Level       & Hierarchy                         &                                   \\
\midrule
Chip      & Grid        & HBM                               & 128 GB / GPU                      \\
\addlinespace 
\multirow{2}{*}{XeCore}    & CTA   & L1 Cache                          & 512 KB / XeCore                   \\
      & (Workgroup) & SLM                               & 128 KB / XeCore                   \\
\addlinespace 
EU        & Warp        & \multirow{2}{*}{Register File}    & \multirow{2}{*}{512 KB / XeCore}  \\
Lane      & Thread      &                                   &                                   \\
\bottomrule
\end{tabularx}
\end{table}

\subsection{Triton}
A high-performance kernel is essential for maximizing the computational power of a GPU, thereby accelerating both training and inference processes. Traditionally, handwritten vendor libraries such as CUTLASS~\cite{cutlass} have been the go-to solutions for achieving optimal performance. However, these approaches often demand a deep understanding of hardware intricacies and lack the flexibility required by researchers who frequently experiment with novel ideas and seek solutions that deliver good performance out-of-the-box.

Triton is emerging as a new programming language tailored specifically for GPU kernel development. With its Python-like syntax and workgroup level programming interface, Triton makes GPU programming more accessible to AI researchers and engineers. It allows researchers to focus on algorithmic innovation and high-level optimizations without being bogged down by low-level hardware details. 
%Underneath, triton compiler optimizes the code to match the performance of native framework operators.

Currently, Triton is the default backend for TorchInductor~\cite{pytorch2}, enabling PyTorch ATen operators to be dispatched to pre-written Triton template kernels. TorchInductor also supports Just-In-Time (JIT) generation of element-wise and reduction operations in Triton, which can then be fused with Triton template kernels. This approach provides greater flexibility compared to traditional pre-defined operator fusion patterns, allowing for more dynamic and efficient execution.

Moreover, an increasing number of frameworks and tools are adopting Triton for kernel development, including vLLM~\cite{vllm}, Mamba~\cite{gu2023mamba} and DeepSpeed~\cite{deepspeed}.

\subsubsection{Triton Dialect}
\hfill\\
The latest Triton compiler is built on top of MLIR~\cite{lattner2020mlircompilerinfrastructureend}, leveraging its extensive set of built-in utilities. Besides reusing existing dialects such as arith, math, and scf for computation and control flow, Triton introduces its own dialect, known as the Triton dialect (abbreviated as tt), specifically designed to express block-level operations on tensors.

In Triton, tensor represents an N-dimensional array of either values or pointers. By default, Triton uses tensor of pointers as the primary mechanism for memory access. This means that each element in the tensor is a pointer, representing a block of pointers. Later Triton introduced pointer to a block tensor (block pointer) to represent a contiguous block of data. However, in the default compilation pipeline, all block pointers are eventually rewritten into tensors of pointers. While this approach is general enough to handle sparse operations, it necessitates heavy memory analysis to determine data contiguity. For dense operations, we argue that using block pointer is a more efficient approach because it explicitly conveys contiguity information.

Table~\ref{tab:tritonops} outlines the primary operations associated with the Triton dialect that will be covered in the following chapters. For more detailed information, please refer to the Triton Dialect definition~\cite{tritonops}.

\begin{table}
  \caption{Triton Dialect}
  \label{tab:tritonops}
\begin{tabularx}{\linewidth}{ll}
\toprule
Operations &  Description                                      \\
\midrule
get\_program\_id  & get ID of the current program/workgroup         \\
load              & load a tensor from pointer                      \\
store             & store a tensor to pointer                       \\
dot               & matrix multiplication                           \\
reduce            & reduce along tensor's specified axis            \\
make\_tensor\_ptr & returns a pointer to a block in a tensor        \\
advance           & advance offsets of the tensor pointer           \\
\bottomrule
\end{tabularx}
\end{table}

\subsubsection{Layout Encoding}
\hfill\\
 A tensor's RankedTensorType includes a default \emph{encoding}~\cite{encoding} attribute that can provide additional information to the tensor. Triton takes advantage of this and introduces \emph{layout encoding} with careful design. The layout encoding indicates how data should be partitioned across threads~\cite{tritonencoding}. 
%%Formally speaking, we define a layout as a function \mathcal{L} that maps a multi-dimensional tensor index $i \in \mathbb{Z}^d$ to a set of integers T corresponding to the indices of the threads allowed to access some data at index $i$.
Listing~\ref{code-layout} shows the layout encoding that will be used.

\textbf{BlockedEncoding} represents a contiguous portion of a tensor. The parameters for this encoding include:
\begin{itemize}
\item \emph{sizePerThread}: Specifies the block size that each thread operates on.
\item \emph{threadsPerWarp}: Defines the arrangement of threads within a warp
\item \emph{warpsPerCTA}: Defines the arrangement of warps within a CTA.
\item \emph{order}: Determines the memory access order, with the fastest-changing axis first.
\end{itemize}

\begin{figure}[htb]
    \includegraphics[scale=0.4]{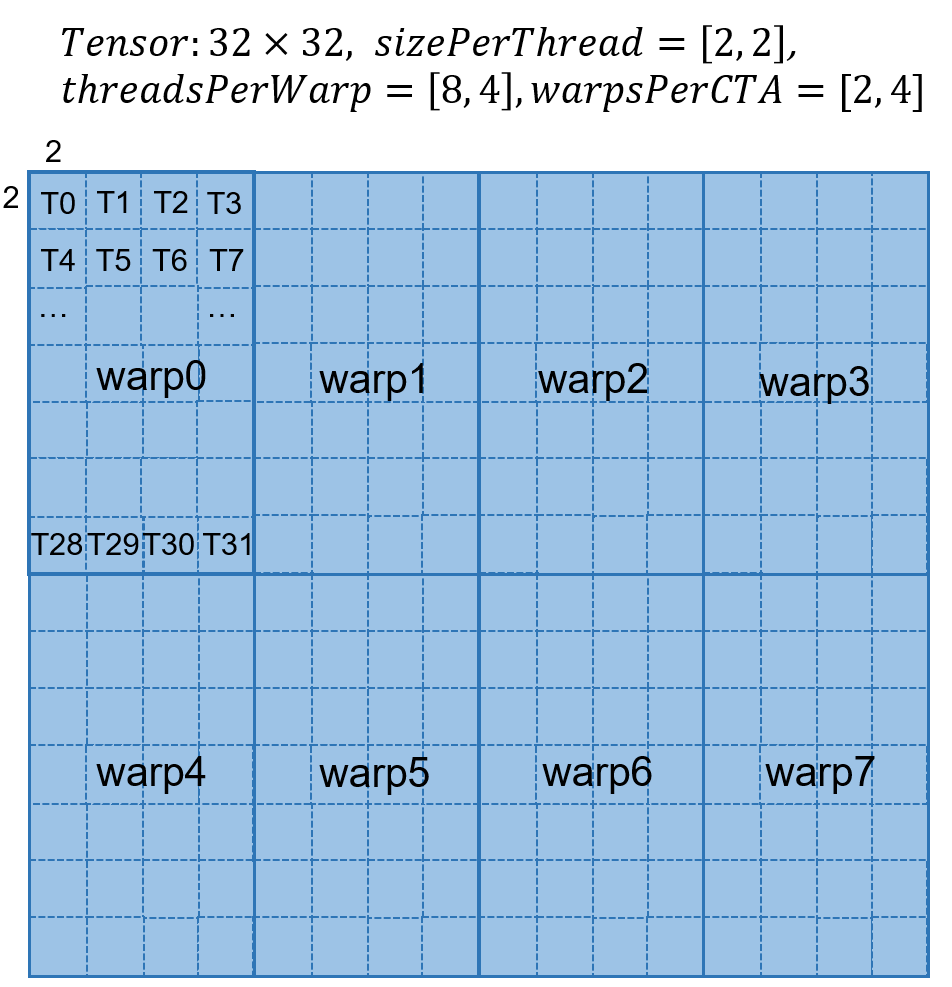}
    \caption{Triton BlockedEncoding}
    \label{fig:blockedEncoding}
\end{figure}

Figure ~\ref{fig:blockedEncoding} provides a visual representation of a typical BlockedEncoding in Triton. In this example, each thread processes a 2x2 block, 4 threads in a row and 8 threads in a column form a warp, 4 warps in a row and 2 warps in a column form a CTA. As a result, each warp handles a 16x8 block, each CTA works on a 32x32 block.

\textbf{DotOperandEncoding} is used for operands in a dot operation.
Take \emph{d = tt.dot a, b, c} for example, both \emph{c} and \emph{d} share the same layout encoding, \emph{a} and \emph{b} have DotOperandEncoding with their \emph{parent} being \emph{c}'s layout encoding, \emph{a}'s \emph{opIdx} is 0, \emph{b}'s \emph{opIdx} is 1, indicating their respective positions in the dot operation.

\textbf{SliceEncoding} indicates its layout is squeezed along the \emph{dim} dimension of the \emph{parent} layout encoding. Take \emph{dst = tt.reduce src, dim} for example, \emph{dst} has a SliceEncoding with its \emph{parent} being \emph{src}'s layout encoding and \emph{dim} being the dimension to be reduced.

%{lexer.py:MlirLexer -x}
\begin{listing}
\caption{Triton layout encoding}
\label{code-layout}
\begin{minted}
[framesep=1.5mm, 
frame=lines,
baselinestretch=1.1,
fontsize=\footnotesize,
escapeinside=||,
breaklines=true,
]{LLVM}
// BlockedEncoding
#blocked = #triton_gpu.blocked<{sizePerThread = [2, 2], threadsPerWarp = [8, 4], warpsPerCTA = [1, 2], order = [1, 0]}>
// DotOperandEncoding
#dot = #triton_gpu.dot_op<{opIdx = 1, parent = #blocked}>
// SliceEncoding
#slice = #triton_gpu.slice<{dim = 1, parent = #blocked}>
\end{minted}
\end{listing}

\subsubsection{Compilation Flow}
\begin{figure}[htb]
    \includegraphics[scale=0.40]{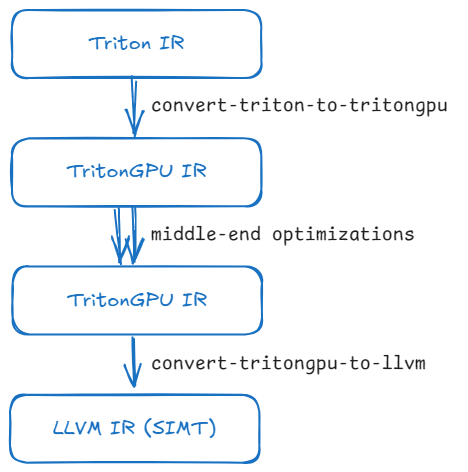}
    \caption{Triton compilation flow}
    \label{old_flow}
\end{figure}

\hfill\\
In Triton, kernel functions are decorated with \emph{triton.jit}. Triton compiler will first walk the Abstract Syntax Tree (AST) of the kernel function to generate Triton IR on-the-fly using a standard SSA construction algorithm~\cite{tritonintro}. Later, Triton IR is converted to Triton GPU IR by adding a naive layout encoding to each tensor type. The Triton GPU IR then undergoes a series of middle-end optimizations aimed at analyzing and simplifying the code. These optimizations include memory coalescing, dot-product-specific enhancements, software pipelining, etc. Finally, the Triton GPU IR is converted to LLVM IR which can then be passed to GPU backend compiler to generate binary for execution. 
Figure ~\ref{old_flow} illustrates the compilation flow. 
%Triton IR consists of a set of Triton operations, alongside other MLIR operations(arith, math, scf). Triton GPU IR retains the same operations as Triton IR, but with each tensor type annotated with a layout encoding to specify how the data is partitioned.
%exactly

\section{Compilation Flow}

\begin{figure}[htb]
    \includegraphics[scale=0.5]{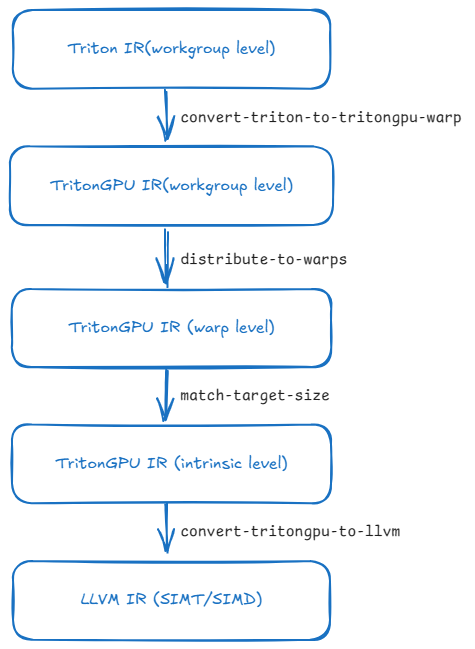}
    \caption{proposed compilation flow}
    \label{fig:flow}
\end{figure}
%%starting from the workgroup level Triton IR
Our proposed compilation flow, illustrated in Figure~\ref{fig:flow}, implements a multi-level lowering that reflects the GPU hierarchy. This approach decouples considerations at different layers, allowing for more efficient and targeted optimizations.

Initially, Triton IR operates at the \textbf{workgroup level}, then we convert it to TritonGPU IR by adding appropriate layout encoding to specify its data distribution between warps. The following \textit{distribute-to-warps} pass will transform the kernel workload to \textbf{warp level} i.e. what each warp should work on. The \emph{match-target-size} pass further split operations to match the LLVM intrinsic size that vendor target can support which we refer to as the \textbf{intrinsic level}. Finally, the TritonGPU IR is converted to LLVM IR with either SIMT or SIMD style.

Next, we will use a GEMM example~\cite{gemm_triton_kernel} to illustrate the compilation flow, as GEMM is a typical AI workload people are most familiar with.

Given $A \in \mathbb{R}^{m \times k} \: B\in \mathbb{R}^{k \times n} \: C\in \mathbb{R}^{m \times n}$, the GEMM is $C \mathrel{+}= A * B$. User can configure the triton kernel to process a workload of  $256x256 \mathrel{+}= 256x32 * 32x256$ in the loop body, with the number of warps (numWarps) set to 32. Listing~\ref{code-ttg} presents the Triton IR after parsing the AST. 

%{lexer.py:MlirLexer -x}
\begin{listing*}
\caption{GEMM Triton IR (w/o highlighted layout encoding) \\
GEMM TritonGPU IR (w/ highlighted layout encoding)}
\label{code-ttg}
\begin{minted}[
framesep=1.5mm, 
frame=lines,
baselinestretch=1.1,
fontsize=\footnotesize,
escapeinside=||,
highlightcolor=my,
highlightlines=(1-3)
]{LLVM}
#blocked = #triton_gpu.blocked<{sizePerWarp = [32, 64], warpsPerCTA = [8, 4], order = [1, 0]}>
#dot0 = #triton_gpu.dot_op<{opIdx = 0, parent = #blocked}>
#dot1 = #triton_gpu.dot_op<{opIdx = 1, parent = #blocked}>
tt.func public @matmul_kernel_with_block_pointers(%arg0: !tt.ptr<f16>, %arg1: !tt.ptr<f16>, %arg2: !tt.ptr<f32>) {
  %cst = arith.constant dense<0.000000e+00> : tensor<256x256xf32, |\colorbox{my}{#blocked}|>
  %a_ptr = tt.make_tensor_ptr %arg0, [%c4096, %c4096], [%c4096, %c1], [%offsetY, %c0] : <tensor<256x32xf16, |\colorbox{my}{#dot0}|>>
  %b_ptr = tt.make_tensor_ptr %arg1, [%c4096, %c4096], [%c4096, %c1], [%c0, %offsetX] : <tensor<32x256xf16, |\colorbox{my}{#dot1}|>>
  %loop:3 = scf.for %arg3 = %c0 to %c4096 step %c32 iter_args(%c = %cst, %arg5 = %a_ptr, %arg6 = %b_ptr) ... {
    %a = tt.load %arg5 : !tt.ptr<tensor<256x32xf16, |\colorbox{my}{#dot0}|>>
    %b = tt.load %arg6 : !tt.ptr<tensor<32x256xf16, |\colorbox{my}{#dot1}|>>
    %d = tt.dot %a, %b, %c : tensor<256x32xf16, |\colorbox{my}{#dot0}|> * tensor<32x256xf16, |\colorbox{my}{#dot1}|> -> tensor<256x256xf32, |\colorbox{my}{#blocked}|>
  }
  %d_ptr = tt.make_tensor_ptr %arg2, [%c4096, %c4096], [%c4096, %c1], [%offsetY, %offsetX] : <tensor<256x256xf32, |\colorbox{my}{#blocked}|>>
  tt.store %d_ptr, %loop#0 : !tt.ptr<tensor<256x256xf32, |\colorbox{my}{#blocked}|>>
}
\end{minted}
\end{listing*}

\subsection{Convert-triton-to-tritongpu-warp}

This pass begins by analyzing the kernel’s workload pattern (e.g., element-wise, reduction, gemm, attention) and then figures out the optimal layout encoding for root operation such as \emph{tt.dot} operation. Subsequently, we get all other value’s layout encoding through def-use chain propagation. 

We make the propagation rules straight-forward: apart from the rules for \emph{tt.dot} and \emph{tt.reduce} introduced earlier, other operations- including \emph{tt.load, tt.store, tt.advance and arith/math.unary/binary operations} require that all source operands and results share the same layout encoding.
%% remove tt.broadcast

There are three major differences from triton upstream.
\begin{enumerate}
    \item \textbf{Workload-Aware}. A dot operation may require different partition strategies depending on the workload to achieve optimal performance. For instance, a square partition is most beneficial for typical GEMM, while FlashAttention-2~\cite{dao2023flashattention2fasterattentionbetter} prefers a partition along the row dimension to minimize inter-warp communication and achieve peak performance.
    \item \textbf{One-Off layout encoding}: Our approach determines the layout encoding in a single step, whereas the upstream triton initially assigns a naive layout encoding and refines it in subsequent passes.
    \item \textbf{Focus on sizePerWarp}: We aim to get \textit{sizePerWarp} (the block size per warp works on) rather than \textit{sizePerThread}. It would be a pre-mature lowering to get what each thread works on at the beginning.
\end{enumerate}	

So, for the GEMM example, firstly we need to figure out the layout encoding for the root operation - \emph{c += tt.dot a, b}. Given that 
\begin{minted}[escapeinside=||]{llvm}
|c's| workgroupSize = [256, 256], numWarps = 32
\end{minted} 
By applying square partitioning between warps , we get \emph{c}'s \emph{BlockedEncoding}:
\begin{minted}{llvm}
warpsPerCTA = [8, 4]
sizePerWarp = [workgroupSize / warpsPerCTA] = [32, 64]
\end{minted}
Then \emph{a} and \emph{b} have DotOperandEncoding respectively. Finally by propagation rule described above, all tensor types are annotated with a layout encoding as shown in Listing~\ref{code-ttg}. 
Note that our generated Triton GPU IR retains the same operations as Triton IR. \footnote{Upstream triton would introduce many "convert-layout" operations to help the lowering work.}

\subsection{Distribute-to-warps}
This pass distributes the workload of a workgroup across warps according to the corresponding layout encoding. After the pass, we get what each warp works on. 
Previously we modify BlockedEncoding to include \emph{sizePerWarp} and \emph{WarpsPerCTA}—these parameters determine how the workload is distributed.

So, the first step is to get the equivalent BlockedEncoding for every layout encoding. For DotOperandEncoding and SliceEncoding, we derive from its parent layout encoding. The mapping rules are detailed in Table~\ref{tab:layout_transform}. 

\begin{table*}[htb]
  \caption{Layout Encoding Mapping Rule}
  \label{tab:layout_transform}
\begin{NiceTabular}{ll}
\toprule
Original Layout Encoding & Equivalent BlockedEncoding      \\
\midrule
\#blocked: sizePerWarp = pS, warpsPerCTA = pW\tabularnote{\footnotesize\textit{pS} and \textit{pW} are Integer Arrays} & NA               \\
\#dot0 = \#triton\_gpu.dot\_op<\{opIdx = 0, parent = \#blocked\}> & sizePerWarp = [pS[0], shape\tabularnote{\footnotesize{shape} is the static size of the RankedTensorType to which this layout encoding is attached}, warpsPerCTA = pW \\
\#dot1 = \#triton\_gpu.dot\_op<\{opIdx = 1, parent = \#blocked\}> & sizePerWarp = [shape[0], pS[1]], warpsPerCTA = pW \\
\#slice = \#triton\_gpu.slice<\{dim = i, parent = \#blocked\}>    & sizePerWarp = pS.erase(i), warpsPerCTA = pW.erase(i) \\
\bottomrule
\end{NiceTabular}
\end{table*}

\begin{figure}[htb]
    \includegraphics[scale=0.6]{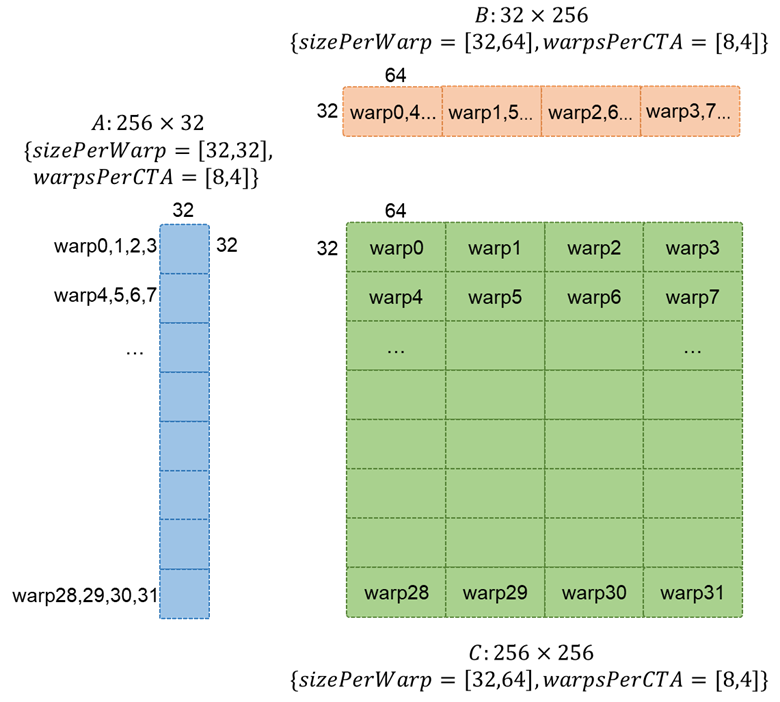}
    \caption{GEMM warp distribution}
    \label{fig:distribute-to-warps}
\end{figure}

%{lexer.py:MlirLexer -x}
\begin{listing*}[htb]
\caption{GEMM TritonGPU IR after distribute-to-warps}
\label{code-distribute}
\begin{minted}
[
frame=lines,
framesep=2mm,
baselinestretch=1.1,
fontsize=\footnotesize,
escapeinside=||,
highlightcolor=my,
]{llvm}
tt.func public @matmul_kernel_with_block_pointers(%arg0: !tt.ptr<f16>, %arg1: !tt.ptr<f16>, %arg2: !tt.ptr<f32>) {
  %cst = arith.constant dense<0.000000e+00> : tensor<32x64xf32, #blocked>
  %warp_id = gpu.subgroup_id : index
  ...
  %a_ptr = tt.make_tensor_ptr %arg0, [%c4096, %c4096], [%c4096, %c1], [%offsetY, %c0] : <tensor<32x32xf16, #dot0>>
  %b_ptr = tt.make_tensor_ptr %arg1, [%c4096, %c4096], [%c4096, %c1], [%c0, %offsetX] : <tensor<32x64xf16, #dot1>>
  %loop:3 = scf.for %arg3 = %c0 to %c4096 step %c32 iter_args(%c = %cst, %arg5 = %a_ptr, %arg6 = %b_ptr) ... {
    %a = tt.load %arg5 : !tt.ptr<tensor<32x32xf16, #dot0>>
    %b = tt.load %arg6 : !tt.ptr<tensor<32x64xf16, #dot1>>
    %d = tt.dot %a, %b, %c : tensor<32x32xf16, #dot0> * tensor<32x64xf16, #dot1> -> tensor<32x64xf32, #blocked>
  }
  %d_ptr = tt.make_tensor_ptr %arg2, [%c4096, %c4096], [%c4096, %c1], [%offsetY, %offsetX] : <tensor<32x64xf32, #blocked>>
  tt.store %d_ptr, %loop#0 : !tt.ptr<tensor<32x64xf32, #blocked>>
}
\end{minted}
\end{listing*}
For the GEMM example, Figure~\ref{fig:distribute-to-warps} illustrates the data distribution between warps. Matrix C is evenly distributed, with each warp processing a 32x64 block. For matrix A, the second dimension of \emph{sizePerWarp} already matches \emph{workgroupSize}, yet we have 4 warps in a row to arrange, so warp 0-3 work on the same 32x32 sub-block of A. Similarly, warps 0,4,8..28 work on the same 32x64 sub-block of B.

After the pass, as shown in Listing~\ref{code-distribute}, \emph{tt.dot} is transformed from 256x256 = 256x32 * 32x256 to 32x64 = 32x32 * 32x64, \emph{offsets} of \emph{tt.make\_tensor\_ptr} are adjusted from a function of \emph{tt.program\_id} to a function of \emph{tt.program\_id} and \emph{gpu.subgroup\_id} (aka warp\_id).

\subsection{Match-target-size}

This pass splits operations into multiple smaller operations to match the target LLVM intrinsic size. All values sharing the same layout encoding are split consistently unless a specific operation requires a different size, in which case a \emph{tt.extract}\footnote{In practice, tt.extract is moved to triton\_intel\_gpu dialect} operation is introduced to extract a sub-block from the input tensor. Users can specify options such as maximum load size and maximum dot size.

% {lexer.py:MlirLexer -x}
\begin{listing*}[htb]
\caption{GEMM TritonGPU IR after match-target-size}
\label{code-match}
\begin{minted}
[
frame=lines,
framesep=2mm,
baselinestretch=1.1,
fontsize=\footnotesize,
] {LLVM}
tt.func public @matmul_kernel_with_block_pointers(%arg0: !tt.ptr<f16>, %arg1: !tt.ptr<f16>, %arg2: !tt.ptr<f32>) {
  %cst = arith.constant dense<0.000000e+00> : tensor<8x16xf32>
  %a_ptr  = tt.make_tensor_ptr %arg0, [%c4096, %c4096], [%c4096, %c1], [%offsetY, %c0] : <tensor<32x32xf16>>
  %b_ptr0 = tt.make_tensor_ptr %arg1, [%c4096, %c4096], [%c4096, %c1], [%c0, %offsetX] : <tensor<32x32xf16>>
  %b_ptr1 = tt.make_tensor_ptr %arg1, [%c4096, %c4096], [%c4096, %c1], [%c0, %offsetX + %c32] : <tensor<32x32xf16>>
  %loop:4 = scf.for %arg3 = %c0 to %c4096 step %c32 iter_args(%c_sub = %cst, %arg5 = %a_ptr, %arg6 = %b_ptr0, %arg7 = %b_ptr1) ... {
    %a  = tt.load %arg5 : !tt.ptr<tensor<32x32xf16>>
    %b0 = tt.load %arg6 : !tt.ptr<tensor<32x32xf16>>
    %b1 = tt.load %arg7 : !tt.ptr<tensor<32x32xf16>>
    %a_sub0 = tt.extract %a[0] : tensor<32x32xf16> -> tensor<8x16xf16>
    %b_sub0 = tt.extract %b0[0] : tensor<32x32xf16> -> tensor<16x16xf16>
    %accumulate = tt.dot %a_sub0, %b_sub0, %c_sub : tensor<8x16xf16> * tensor<16x16xf16> -> tensor<8x16xf32>
    ... // 32 tt.dot in all }
  %d_ptr0 = tt.make_tensor_ptr %arg2, [%c4096, %c4096], [%c4096, %c1], [%offsetY, %offsetX] : <tensor<8x16xf32>>
  tt.store %d_ptr0, %loop#0 : !tt.ptr<tensor<8x16xf32>>
}
\end{minted}
\end{listing*}

\begin{figure}[htb]
    \includegraphics[scale=0.43]{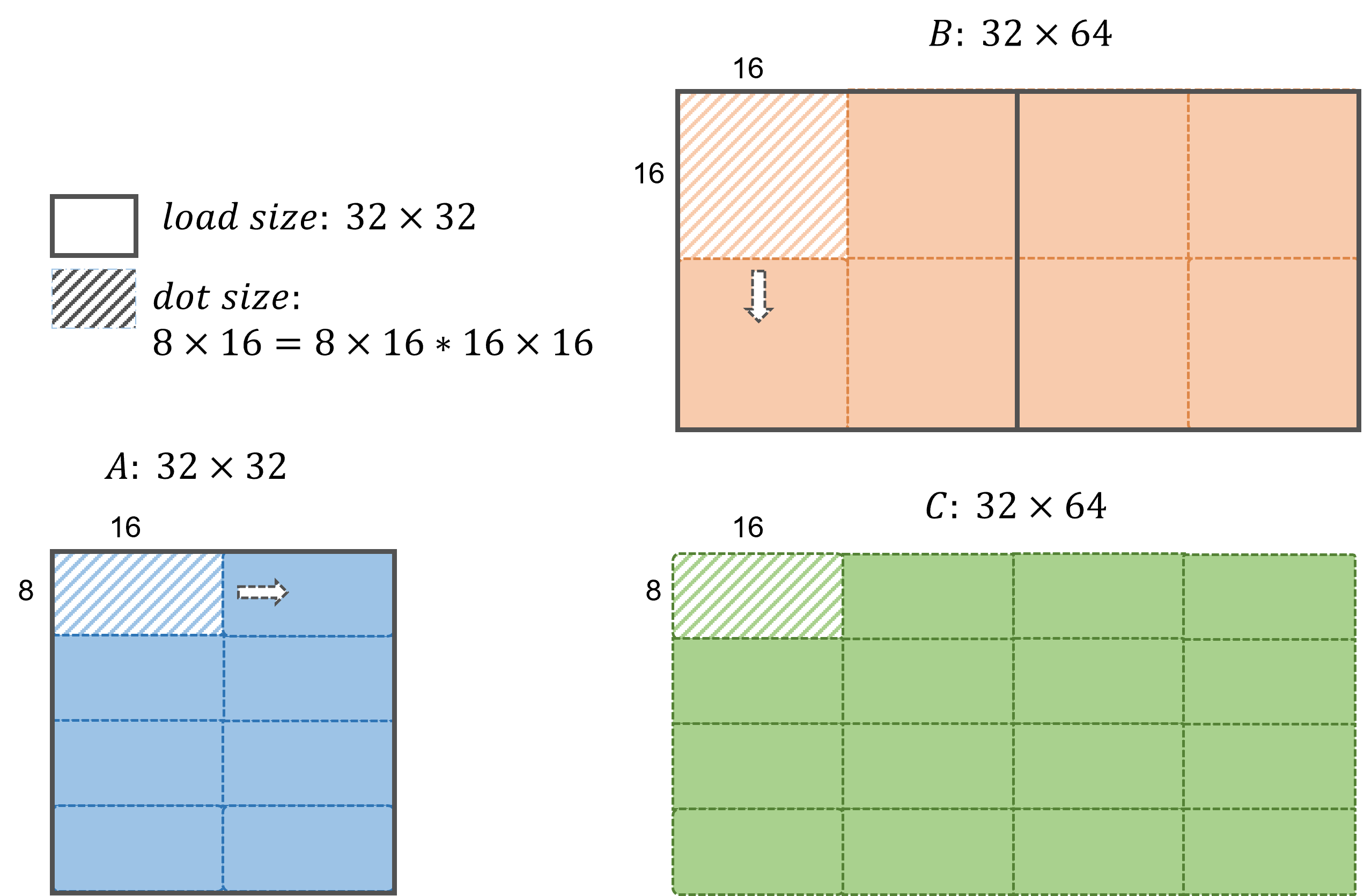}
    \caption{data partition to match target intrinsic size}
    \label{fig:match size}
\end{figure}

For the GEMM example,  PVC’s max load size is 32x32, max dot size is 8x16 = 8x16 * 16x16. The data partitioning is illustrated in Figure~\ref{fig:match size}.

After the pass, as shown in Listing~\ref{code-match}, the load for A with size 32x32 already matches the target load size, so it remains unchanged. However, the load for B with size 32x64 is spilt into 2 load operations. The dot operation, originally 32x64 = 32x32 * 32x64, is split into 32 smaller dot operations of size 
8x16 = 8x16 * 16x16. All values in the def-use chain of the load are spilt into operations working on 32x32 block. Since \emph{tt.dot} requires different block size, \emph{tt.extract} is added to extract a 8x16 sub-block from A and a 16x16  sub-block from B, the sub-blocks are then fed to the dot operation.
\emph{tt.extract} will be lowered to sub-register access in the assembly code, without introducing any register moves.

\subsection{Convert-tritongpu-to-llvm}

This pass converts all operations to LLVM IR. It reuses upstream MLIR conversions for arith, math and scf operations. For triton operations, separate conversion patterns are used to map them to LLVM intrinsic. PVC GPU backend compiler provides two sets of intrinsic: VectorCompute-Intrinsic~\cite{vc_instrisic} for SIMD programming and GenISA-Intrinsic~\cite{genisa_intrinsic} for SIMT programming.

Basically, the conversion is a 1 to 1 mechanical mapping since we already have the operations match the target supported intrinsic size. Only that for SIMT conversion, the data is evenly distributed to each thread lane, meaning the vector size in the intrinsic need to be divided by \emph{threadsPerWarp}. 

\begin{listing*}[htb]
\caption{GEMM LLVM IR}
\label{code-llvm}
\begin{minted}
[
frame=lines,
framesep=2mm,
baselinestretch=1.1,
fontsize=\footnotesize,
]{LLVM}
;SIMD style
%a  = call <512 x i32> @llvm.genx.lsc.load2d.stateless.v512i32.i1.i64(..., i64 %a_ptr, ...)
%b0 = call <512 x i32> @llvm.genx.lsc.load2d.stateless.v512i32.i1.i64(..., i64 %b_ptr0, ...)
%b1 = call <512 x i32> @llvm.genx.lsc.load2d.stateless.v512i32.i1.i64(..., i64 %b_ptr1, ...)
%a_sub0 = shufflevector <512 x i32> %a, <512 x i32> poison, <64 x i32> <i32 0..63>
%b_sub0 = shufflevector <512 x i32> %b0, <512 x i32> poison, <128 x i32> <i32 0..127>
%d = call <128 x float> @llvm.genx.dpas2.v128f32.v128i32.v64i32(%c_sub, <128 x i32> %b_sub0, <64 x i32> %a_sub0, ...)
;SIMT style
%a  = call <64 x i16> @llvm.genx.GenISA.LSC2DBlockRead.v64i16(i64 %a_ptr, ...)
%b0 = call <32 x i32> @llvm.genx.GenISA.LSC2DBlockRead.v32i32(i64 %b_ptr0, ...)
%b1 = call <32 x i32> @llvm.genx.GenISA.LSC2DBlockRead.v32i32(i64 %b_ptr1, ...)
%a_sub0 = shufflevector <64 x i16> %a, <64 x i16> undef, <8 x i32> <i32 0, i32 1, i32 2, i32 3, i32 4, i32 5, i32 6, i32 7>
%b_sub0 = shufflevector <32 x i32> %b0, <32 x i32> undef, <8 x i32> <i32 0, i32 1, i32 2, i32 3, i32 4, i32 5, i32 6, i32 7>
%d = call <8 x float> @llvm.genx.GenISA.sub.group.dpas.v8f32.v8i16.v8i32(%c_sub, <8 x i16> %a_sub0, <8 x i32> %b_sub0, ...)
\end{minted}
\end{listing*}

\begin{table}[htb]
  \caption{Triton to LLVM}
  \label{tab:tritonllvm}
  \small
  \begin{tabularx}{\linewidth}{lll}
    \toprule
    Triton Ops & LLVM Ops - SIMT & LLVM Ops - SIMD \\
    \midrule
    tt.load A & 2DBlockRead.v64i16 & load2d.stateless.v512i32 \\
    tt.load B & 2DBlockRead.v32i32 & load2d.stateless.v512i32 \\
    tt.dot    & dpas.v8f32.v8i16.v8i32 & dpas2.v128f32.v128i32.v64i32 \\
    tt.store  & 2DBlockWrite.v8i32 & store2d.stateless.v128i32 \\
    \bottomrule
\multicolumn{3}{l}{\small *due to intrinsic constraint, i16/i32 are used instead of f16/f32} \\
  \end{tabularx}
\end{table}

Table~\ref{tab:tritonllvm} shows the conversion. Take \emph{tt.load A} for example, the workload size \emph{32x32xf16} is flattened to \emph{v512i32}, when divided by PVC's threadsPerWarp(16), results in \emph{v64i16}. Listing~\ref{code-llvm} shows the LLVM IR after the conversion.

\subsection{FlashAttention-2}

As demonstrated in the compilation flow, each value’s layout encoding is the key.  The layout encoding dictates how work is distributed among warps and serves as a guide for how each operation should be split to match the target intrinsic size.

Once each value is annotated with the correct layout encoding, the subsequent passes can be applied effectively. And once we figure out the root operation’s layout encoding, the encoding for all other values can be inferred by tracing the def-use chain. This approach naturally facilitates pre- and post-operation fusion, as they can be seamlessly expanded from the root.

Let’s take Flash attention-2~\cite{fa_triton_kernel} as another example. It can be seen as a fused kernel of two back-to-back GEMMs with an online softmax\cite{online-softmax} in between, as outlined in Algorithm~\ref{algo-fa2}.
%%It is commonly used in LLM, compared to standard attention, reduction in shared memory reads/writes yields speedup.

\begin{figure}[htb]
    \includegraphics[scale=0.35]{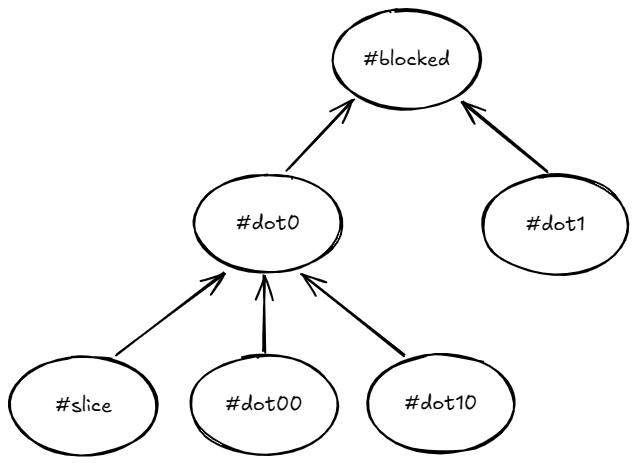}
    \caption{relation of FlashAttention-2 layout encoding}
    \label{fig:tree-fa2}
\end{figure}

We adopt the work partitioning from the original paper~\cite{dao2023flashattention2fasterattentionbetter} which distributes output matrix \emph{O} along the row dimension across all warps and \emph{K/V} accessed by all warps. Balancing the data shared among warps and register pressure, we arrive at the following kernel configuration for PVC:
\begin{minted}[escapeinside=||,]{llvm}
O's workgroupSize = [128, 64], numWarps = 8
\end{minted}
By horizontally partitioning between warps, we get \emph{O}'s \emph{BlockedEncoding}:
\begin{minted}{llvm}
warpsPerCTA = [numWarps, 1] = [8, 1]
sizePerWarp = [workgroupSize / warpsPerCTA] = [16, 64]
\end{minted}
Subsequently, the layout encoding for all other tensors is inferred, as summarized in Table~\ref{tab:tritonfademo}.
Figure~\ref{fig:tree-fa2} illustrates their relationships.

\IncMargin{1em}
\begin{algorithm}
    \DontPrintSemicolon
    \SetAlgoLined
    \SetKwInOut{Input}{input}
    \SetKwFunction{onlinesoftmax}{online\_softmax}
    \SetKwFunction{rowmax}{rowmax}
    \SetKwFunction{rowsum}{rowsum}
    \Input{$Q$, $K$, $V$ $\in \mathbb{R}^{N \times D}$, N is the sequence length, D is the head dimension}
    
    $J \leftarrow N / BlockSize$ \;
    Load $Q$ from HBM\;
    \For{$j$ in $(0, J)$}{
        Load $K_j$ from HBM\;
        Compute $QK = Q*K_j$ \;
        Compute $P$ = \onlinesoftmax $QK$ (intermediate result: $m$ = \rowmax, $l$ = \rowsum) \;
        Load $V_j$ from HBM\;
        Compute $O += P * V_j $ \;
     }
    Store $O$ back to HBM.
    \caption{FlashAttention-2 forward}
    \label{algo-fa2}
\end{algorithm}
\DecMargin{1em}

\begin{table}[htb]
\caption{FlashAttention-2 Layout Encoding}
\label{tab:tritonfademo}
\begin{tabular}{cl}
\toprule
Symol & Layout Encoding                                                                                                                         \\
\midrule
O     & \begin{tabular}[c]{@{}l@{}}\textbf{\#blocked}: sizePerWarp = {[}16, 64{]},\\ 
        warpsPerCTA = {[}8, 1{]}\end{tabular}                               \\
\myVSpace
V     & \begin{tabular}[c]{@{}l@{}}\textbf{\#dot1}   =   \#triton\_gpu.dot\_op\textless{}\{opIdx = 1, \\ parent = \#blocked\}\textgreater{}\end{tabular} \\
\myVSpace
QK, P & \begin{tabular}[c]{@{}l@{}}\textbf{\#dot0}   =   \#triton\_gpu.dot\_op\textless{}\{opIdx = 0, \\ parent = \#blocked\}\textgreater{}\end{tabular} \\
\myVSpace
K     & \begin{tabular}[c]{@{}l@{}}\textbf{\#dot10} = \#triton\_gpu.dot\_op\textless{}\{opIdx = 1, \\ parent =   \#dot0\}\textgreater{}\end{tabular}     \\
\myVSpace
Q     & \begin{tabular}[c]{@{}l@{}}\textbf{\#dot00} = \#triton\_gpu.dot\_op\textless{}\{opIdx = 0, \\ parent =   \#dot0\}\textgreater{}\end{tabular}     \\
\myVSpace
m, l  & \begin{tabular}[c]{@{}l@{}}\textbf{\#slice} = \#triton\_gpu.slice\textless{}\{dim = 1, \\ parent =   \#dot0\}\textgreater{}\end{tabular}      \\  
\bottomrule
\end{tabular}
\end{table}

\section{Language Extension}
Triton, as a DSL for accelerated computing, is inherently extensible. During our development process, we found that by incorporating simple compiler hint and warp level programming, Triton significantly simplifies GPU programming and accelerates the journey to peak performance.
%Many DL compilers[] would introduce declarative annotations like how to  to make compiler transformations transparent, while this is convenient for compiler guys, it is too much for most of the kernel developers. We want to enable developers focus on the algorithm and major.
The tradeoff is that what is the algorithm workload specific or compiler hackable optimization, we choose to let user have control.

\subsection{Compiler Hint}

In the code generation pipeline, we resort to a workload-aware pass to detect specific patterns and determine the root operation’s tiling partition - layout encoding. Actually this setting is derived from the best-known practice of expert’s kernel tuning experience.  Also, there are instances where researchers may want to manually define the tiling partition. For example, FlashAttention-2~\cite{dao2023flashattention2fasterattentionbetter} explicitly proposes how to partition work between different warps to get optimal performance.

Hence, we provide a compiler hint that allows users to specify the root operation’s tiling partition between warps. Below are the tiling options available for 2D tensor.

\textbf{Horizontal}: Evenly tiles along the first(row) dimension.

\textbf{Vertical}: Evenly tiles along the last(column) dimension.

\textbf{Square}: Tiles to form square sub-blocks.

%%\begin{description}
%%\item[\texttt{Horizontal}:]  Evenly tiling the workload along the first/row dimension.
%%\item[\texttt{Vertical}:] Evenly tiling the workload along the last/column dimension.
%%\item[\texttt{Square}:] Tiling the workload to make each sub-block square.
%%\end{description}

For flash attention-2, merely setting the second dot’s \emph{tiling} to \emph{horizontal} is sufficient, no other source code changes are needed. The compiler can then figure out all values’ layout encoding accordingly.
\begin{minted}
[escapeinside=||,
highlightcolor=my
]{python}
o = tl.dot(p, v, o, tiling="horizontal")
\end{minted}

\subsection{Warp Level API}

Writing kernels at the workgroup level reduces the burden on developers, but performance is highly dependent on compiler, which need time to evolve. Rather than relying solely on compiler-specific optimizations, we believe it is more effective to give developers fine-grained control over their code. For instance, FlashAttention-3~\cite{shah2024flashattention3fastaccurateattention} proposes better warp level management to leverage the latest hardware capabilities. Similarly, many kernel libraries like CUTLASS~\cite{cutlass} offer warp level C++ templates.

Thereby, we introduce a warp level language extension. The key elements are as follows:

\textbf{warp\_level}:  Metadata indicating this a warp level kernel.

\textbf{tl.warp\_id()}: Returns linear ID of the current warp within the workgroup.

\textbf{tl.alloc(shape, data type)}: Allocates and returns a pointer to a block in the SLM with the specified \emph{shape} and \emph{data type}.

\textbf{tl.reduce(..., cross\_warp, dst\_warps)}: Adds keyword parameters for reduction operations (e.g., max, sum).
When \emph{cross\_warp} is set to true, it is a reduction across all warps, otherwise it is a reduction within the current warp.
The \emph{dst\_warps} parameter allows the reduction result to be broadcast only to the specified destination warps. If not set, the result will be broadcast to all warps.

LLM inference often involves long sequences for key-value (KV) pairs. To boost throughput, Flash Decoding~\cite{flashdecoding} proposes splitting the keys and values. For efficient memory management, Paged Attention~\cite{vllm} divides the request's KV cache into blocks. 

Listing~\ref{code-paged0} shows how a paged attention triton kernel could be implemented. While the core algorithm remains similar to FlashAttention-2, the warp distribution differs significantly because the sequence length of the query is typically 1. This makes writing a paged attention kernel relatively easy, but achieving optimal performance out-of-the-box is challenging. The compiler must perform specific analyses and optimizations to enhance performance.

%{lexer_python.py:MlirLexer -x}
\begin{listing}[htb]
\caption{paged attention triton kernel - workgroup level}
\label{code-paged0}
\begin{minted}
[
frame=lines,
framesep=2mm,
baselinestretch=1.1,
fontsize=\footnotesize,
linenos
]{python}
q = tl.load(Q_block_ptr)
for i in range(num_blocks):
    k = tl.load(Ki_block_ptr)
    qk = tl.dot(q, k)
    m_i = tl.max(qk, axis=1)
    p = tl.exp((qk - m_i[:, None]))
    l_i = tl.sum(p, axis = 1)
    p /= l_i[:, None]
    v = tl.load(Vi_block_ptr)
    o += tl.dot(p.to(tl.float16), v)
tl.store(O_block_ptr, o)
\end{minted}
\end{listing}

\begin{listing}[htb]
\caption{paged attention triton kernel - warp level}
\label{code-paged1}
\begin{minted}
[
frame=lines,
framesep=2mm,
baselinestretch=1.1,
fontsize=\footnotesize,
linenos,
highlightcolor=my,
highlightlines={2-6, 11, 14, 19, 20}
]{python}
# warp 0 load Q from HBM and store it to SLM
slm_block_ptr = tl.alloc(shape=(1, D), dtype=tl.float16)
if tl.warp_id() == 0:
    q = tl.load(Q_block_ptr)
    tl.store(slm_block_ptr, q)
tl.barrier()
q = tl.load(slm_block_ptr)
k = tl.load(Ki_block_ptr)
qk = tl.dot(q, k)
m_i = tl.max(qk, axis=1)
m_i = tl.max(m_i, cross_warp = True) # sync partial max
p = tl.exp((qk - m_i[:, None]))
l_i = tl.sum(p, axis = 1)
l_i = tl.sum(l_i, cross_warp = True) # sync partial sum
p /= l_i[:, None]
v = tl.load(Vi_block_ptr)
o = tl.dot(p.to(tl.float16), v)
# reduce the Output to warp 0 and store it back to HBM
o = tl.sum(o, cross_warp = True, dst_warps=(0)) 
if tl.warp_id() == 0:
    tl.store(O_block_ptr, o)
\end{minted}
\end{listing}

\begin{figure}[htb]
    \includegraphics[scale=0.35]{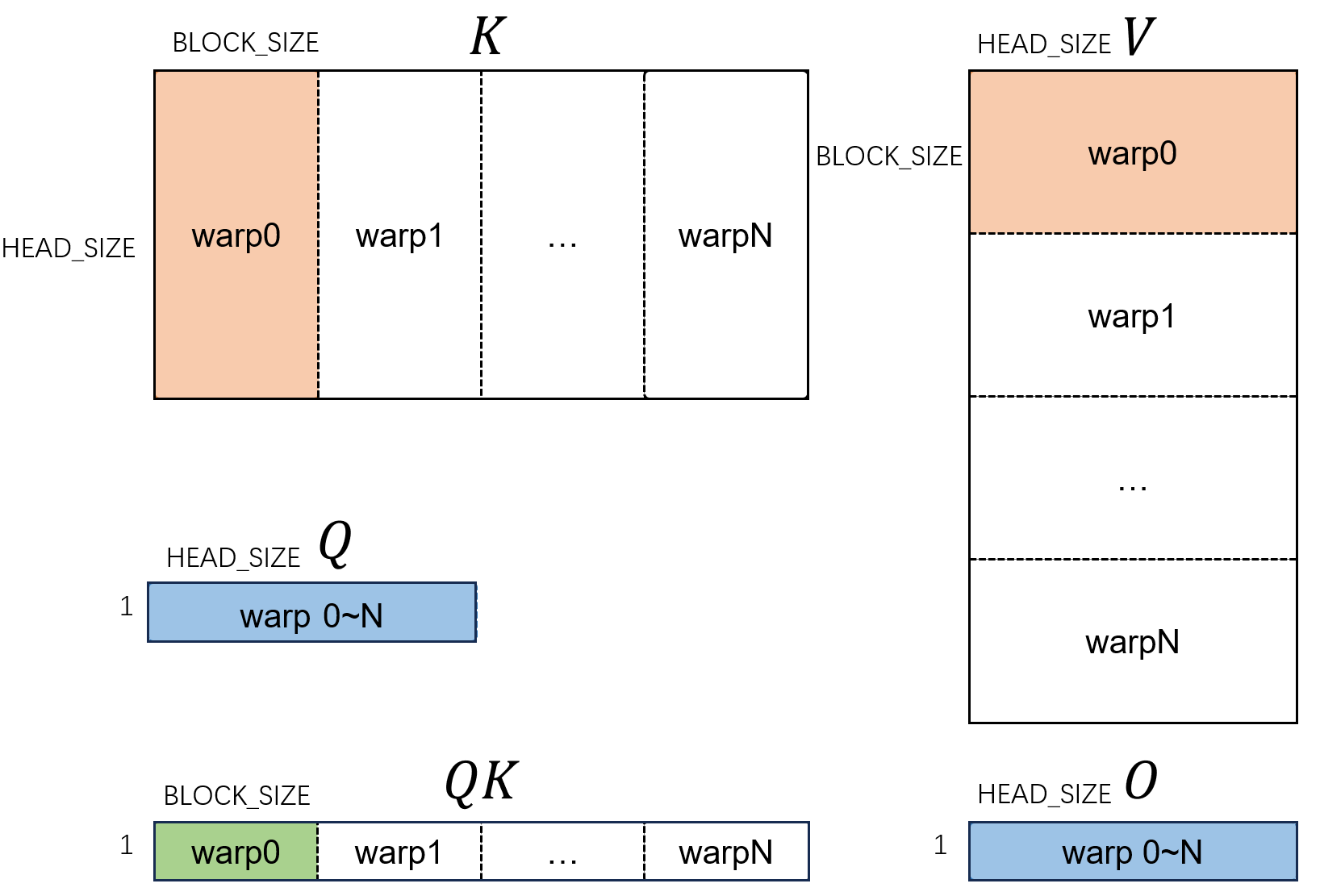}
    \caption{paged attention warp distribution}
    \label{paged}
\end{figure}

To ensure sufficient parallelism on the GPU, one approach is to further partition the KV cache between warps, as depicted in Figure~\ref{paged}. Each warp works on separate KV cache blocks, and reduction across warps is needed to synchronize each warp's partial result.

However, by programming Triton at warp level, users can easily express the above decomposition. Listing~\ref{code-paged1} shows what the warp level kernel would be.

%% user only cares about what each warp's work, compiler will handle it's done by one or multiple intrinsics to accomplish a warp's work,and handles the bother of mapping to thread-lane

\section{Experimental Results}

In this section, we aim to demonstrate the effectiveness of our design by evaluating the performance of several popular AI workload kernels.
% one tile of pvc
 
The experiments were conducted on Intel’s PVC max 1550 using OneAPI 2024.1. 
Performance was measured based on the kernel's GPU execution time recorded by SYCL profiling event ~\cite{sycl}. For comparison, we benchmark Triton against Intel’s XeTLA ~\cite{xetla}, Xe template-based linear algebra library optimized as a peak-performance reference, similar to NVIDIA’s CUTLASS. To ensure a fair comparison, we used identical configurations for both XeTLA and Triton, including parameters such as tile size, minimizing any performance discrepancies due to these settings.

\subsection{GEMM}

GEMM is a fundamental operation in the AI domain, constituting a significant portion of the computational workload. We used the GEMM kernel ~\cite{gemm_triton_kernel} from the Triton tutorial for our tests.

We evaluated two types of GEMM operations: memory-bound and compute-bound. All matrix shapes were derived from LLM models such as LLama-2 and LLama-3.
 
\begin{figure}[htb]
    \includegraphics[scale=0.65]{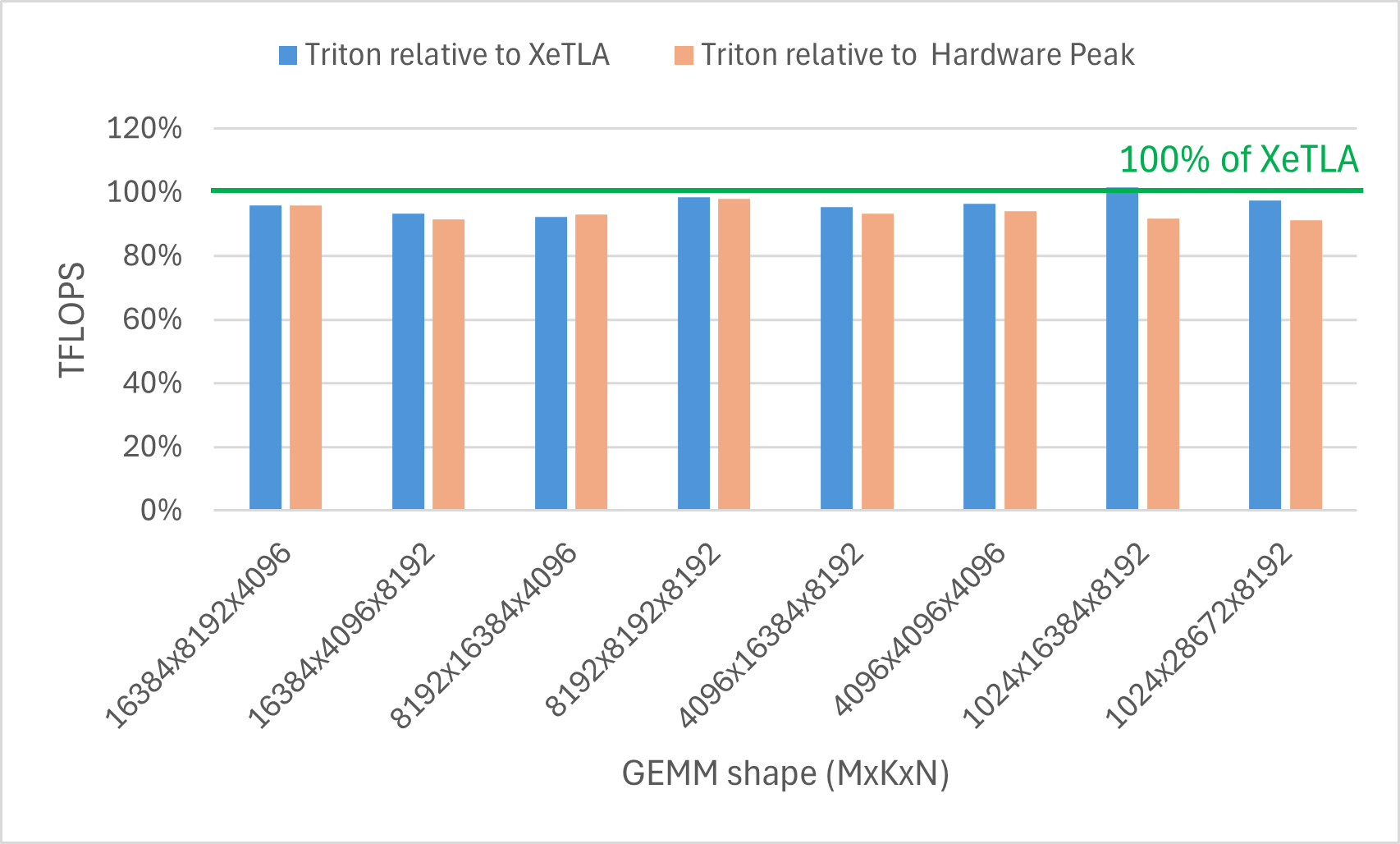}
    \caption{compute-bound GEMM performance}
    \label{fig:perf_gemm_compute}
\end{figure}

Compute-bound GEMM is relevant for both LLM training and inference. As researchers focus on long context length on a single GPU, we tested matrix sizes ranging from m = 1k to 16k, large enough to fully utilize the GPU and achieve peak hardware throughput.  Figure~\ref{fig:perf_gemm_compute} shows that Triton achieves a geometric mean of $96\%$ of XeTLA’s performance.
 
\begin{figure}[htb]
    \includegraphics[scale=0.65]{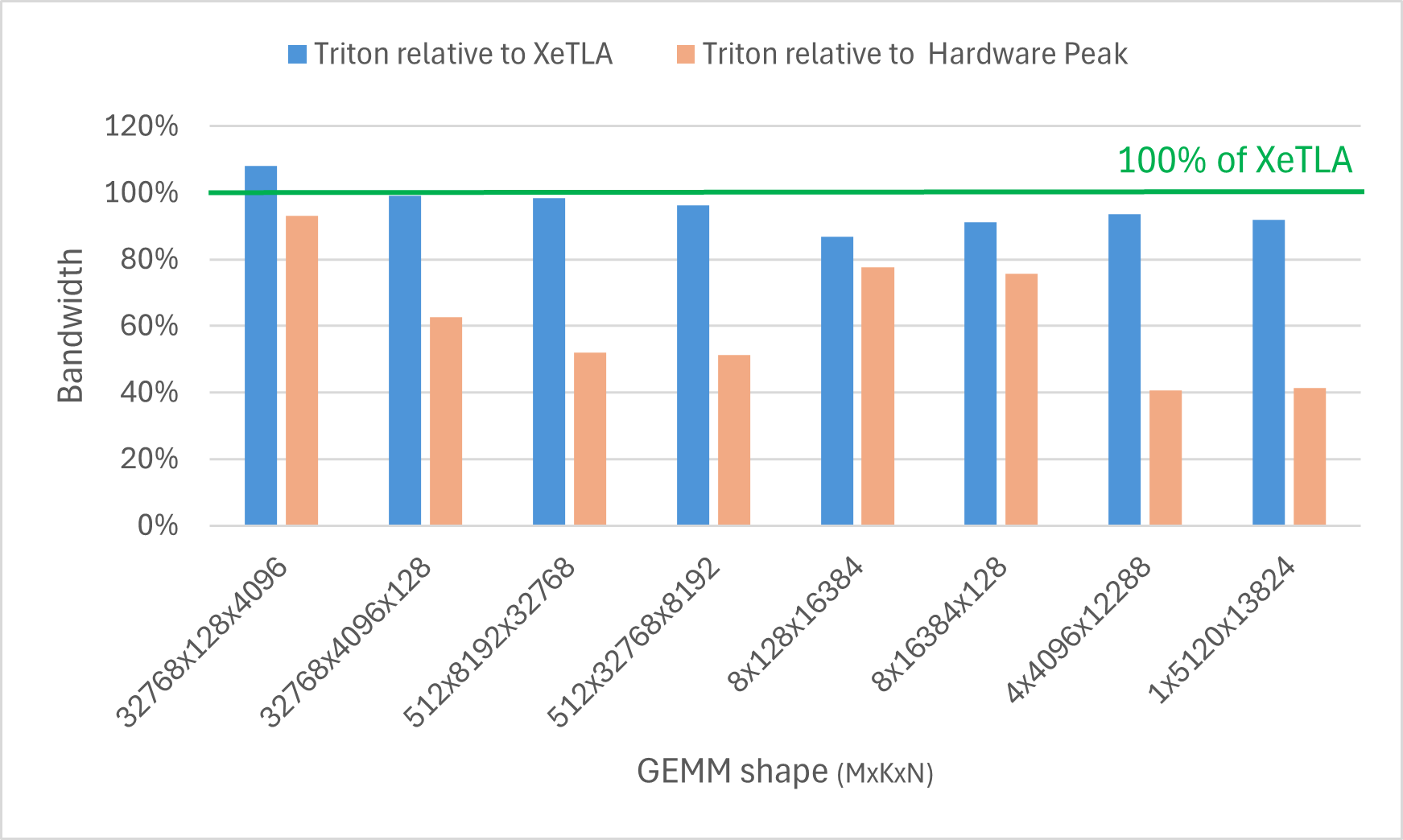}
    \caption{memory-bound GEMM performance}
    \label{fig:perf_gemm_memory}
\end{figure}

Memory-bound GEMM is a common scenario in LLM inference, particularly during the next-token prediction stage. We evaluated this on cases with large m, large k, and large n to demonstrate Triton’s robustness. As shown in Figure~\ref{fig:perf_gemm_memory}, Triton’s performance is comparable to XeTLA, with a $94\%$ geometric mean. 


\subsection{FlashAttention-2}

FlashAttention-2 is widely used for MHA, playing a critical role in modern transformer models. We used the kernel in triton tutorial~\cite{fa_triton_kernel} for our tests.

\begin{figure}[htb]
    \includegraphics[scale=0.65]{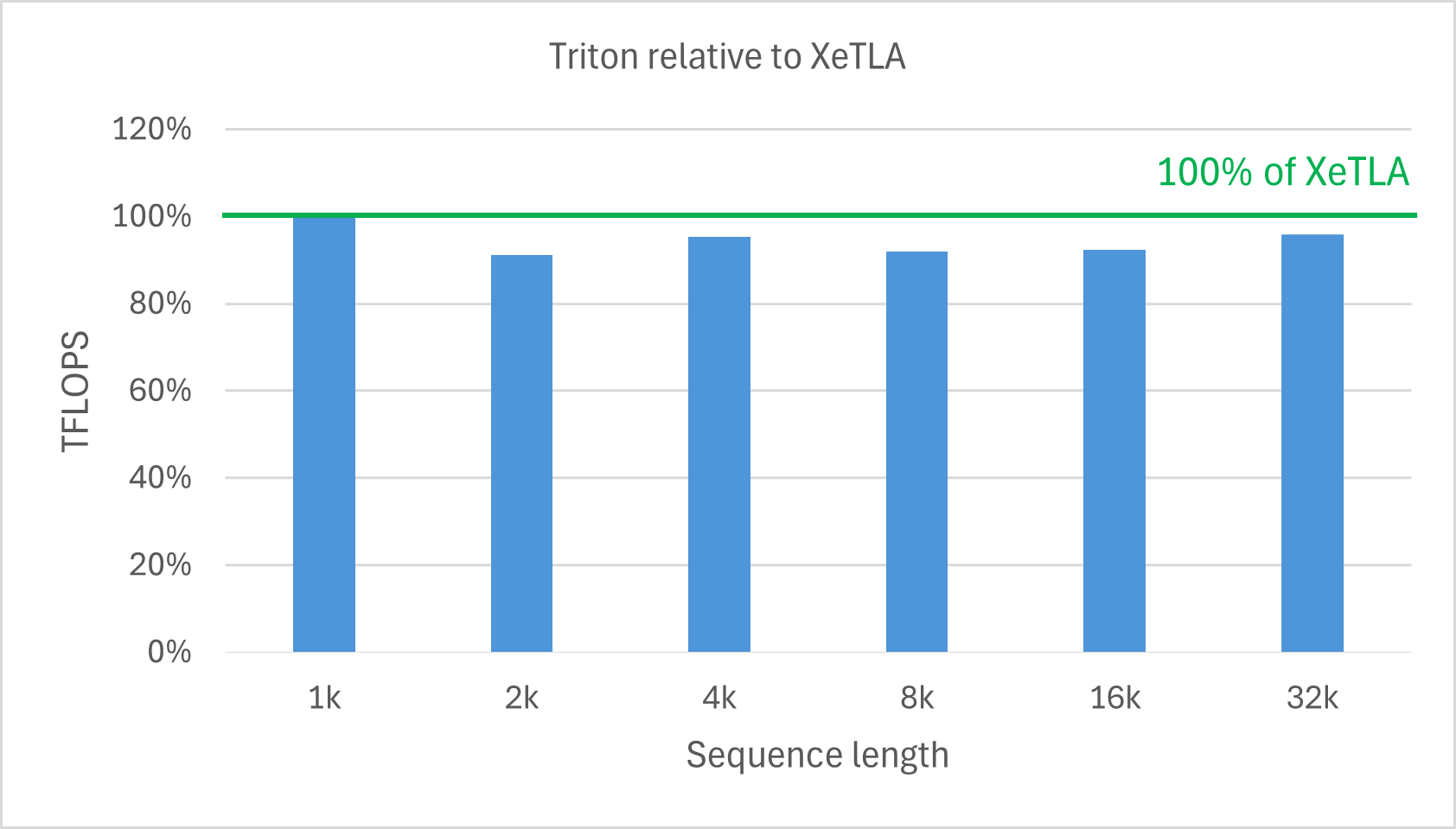}
    \caption[FlashAttention-2 forward performance; head\_dimension = 64]
        {\tabular[t]{@{}l@{}}FlashAttention-2 forward performance \\ head\_dimension = 64\endtabular}
    \label{fig:perf_fa2_64}
\end{figure}

\begin{figure}[htb]
    \includegraphics[scale=0.65]{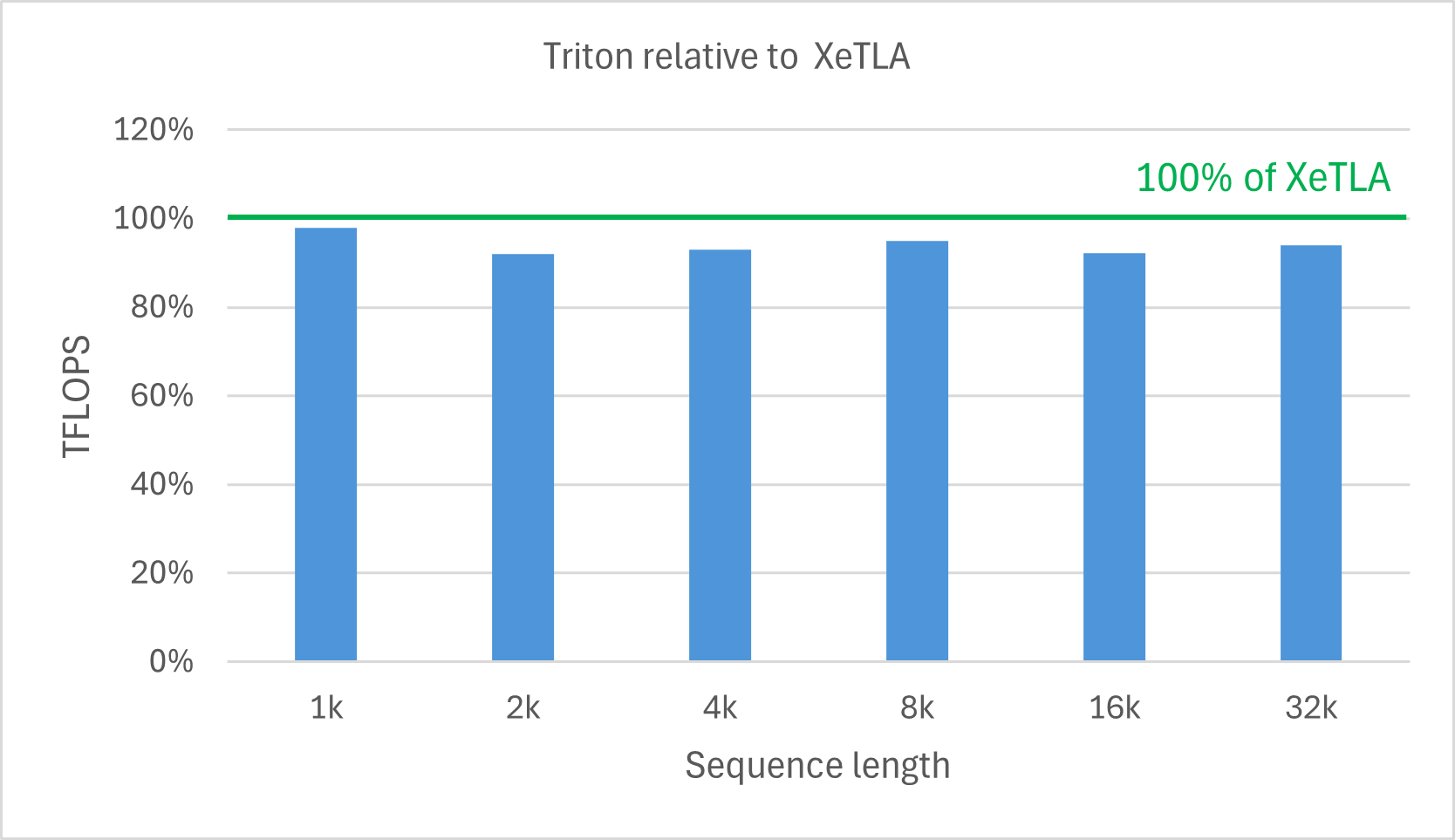}
    \caption[FlashAttention-2 forward performance; head\_dimension = 128]
        {\tabular[t]{@{}l@{}}FlashAttention-2 forward performance \\ head\_dimension = 128\endtabular}
    \label{fig:perf_fa2_128}
\end{figure}

We evaluated the forward pass with a total 32k tokens and sequence length ranging from 1k to 32k, aligned with the context length of most popular LLMs. The hidden dimension was set to 2048, with head dimension to be either 64 or 128 (i.e.,32 heads or 16 heads). The benchmark results in Figure~\ref{fig:perf_fa2_64} and Figure ~\ref{fig:perf_fa2_128} show less than a $5\%$ performance gap, demonstrating the high quality of our code generation.


\subsection{Paged Attention}
Paged attention is widely used in LLM inference engines. Unlike flash attention, the key/value pairs in paged attention are not stored contiguously and must be accessed through a block table mapping, which increases the strain on memory access.
 
Compared to traditional triton workgroup level implementation, our warp level kernel directly express the distribution between warps, requiring only a few additional lines of code.
 
\begin{figure}[htb]
    \includegraphics[scale=0.65]{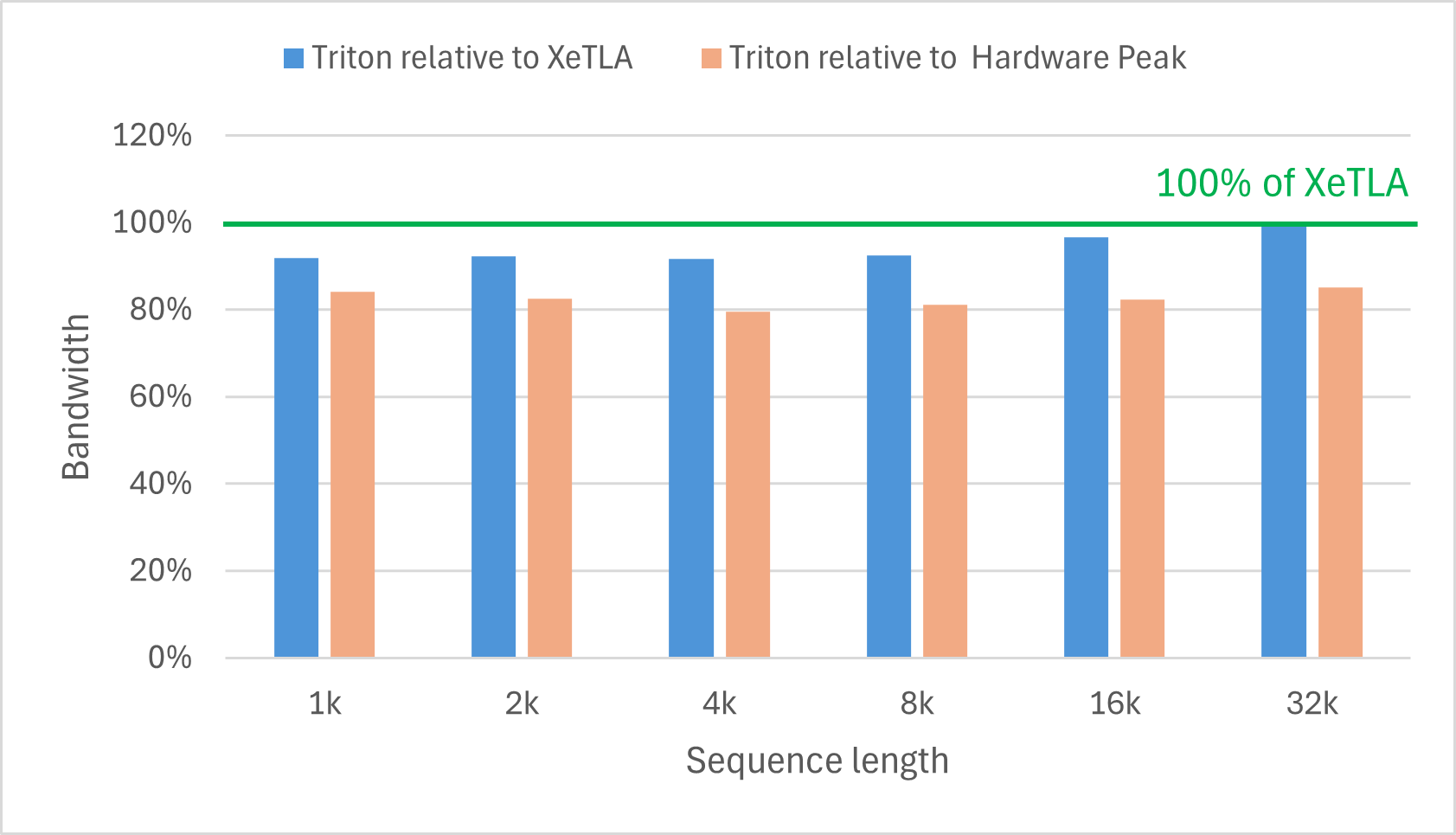}
    \caption{PagedAttention performance}
    \label{fig:perf_pa}
\end{figure}

As shown in Figure~\ref{fig:perf_pa}, Triton’s performance is above $95\%$ of XeTLA's, demonstrating its capability to handle complex kernels effectively.

\section{Conclusion and Future work}

In this paper, we presented ML-Triton which features multi-level lowering and programming interface.
The multi-level compilation flow is closely aligned with the GPU's layered hierarchy. By progressively lowering operations from the workgroup level to the warp level and finally to the intrinsic level, we decompose high-level operations step by step guided by the layout encoding in an innovative straightforward way.

Additionally, we extend Triton language by introducing user-defined compiler hints and warp level programming. These enhancements provide researchers with fine-grained control over their code, reducing dependency on compiler specific optimizations, leading to better out-of-the-box performance.
%allowing the compiler to make informed decisions instead of compiler doing heavy analysis.
%fill the gap between GPU programming efficiency and high performance.

We thoroughly evaluated three popular kernels—GEMM, FlashAttention-2, and Paged Attention—based on our approach, achieving a performance gap of less than $5\%$ compared to expert tuned implementation.

Overall, building on top of Triton, our proposal further bridges the gap between ease of use and high performance in GPU programming.

Looking ahead, we plan to polish our design as we encounter more use cases from the rapidly evolving AI landscape. We also anticipate that this programming and compilation paradigm could be extended beyond GPUs to other many-core architectures.

%as long as appropriate abstraction homogeneous arch.


%\begin{acks}
%  Yudong  Zilan Fangwen Areg Jianhui W\newline
%IREE
%\end{acks}

\newpage
\bibliographystyle{ACM-Reference-Format}
\bibliography{Reference}


\end{document}